\documentclass[letterpaper]{article} 
\usepackage{aaai2027}  
\usepackage[hyphens]{url}  
\usepackage{graphicx} 
\usepackage{natbib}  
\usepackage{caption} 
\usepackage{algorithm}
\usepackage{algorithmic}
\usepackage{amsmath}
\usepackage{multirow}
\usepackage{amssymb}
\usepackage{pifont}
\usepackage{newfloat}
\usepackage{listings}
\DeclareCaptionStyle{ruled}{labelfont=normalfont,labelsep=colon,strut=off} 
\floatstyle{ruled}
\newfloat{listing}{tb}{lst}{}
\floatname{listing}{Listing}

\usepackage{booktabs}
\usepackage{colortbl}
\usepackage{xspace}
\DeclareRobustCommand{\ourmethod}{\textsc{SCORE}\xspace}

\title{Navigating Sparse Evidence: Agentic Visual RAG via\\ Explicit Context Selection and Consolidation}
\author{
    Yucheng Shen\textsuperscript{\rm 1,2}
    Lingyong Yan\textsuperscript{\rm 2}\thanks{Corresponding authors.}, Jiulong Wu\textsuperscript{\rm 2}, Shuaiqiang Wang,\\ Jianmin WU\textsuperscript{\rm 2}, Dawei Yin\textsuperscript{\rm 2},
    Min Cao\textsuperscript{\rm{1}}\footnotemark[1]
}
\affiliations{
    \textsuperscript{\rm{1}}School of Computer Science and
    Technology, Soochow University
    \textsuperscript{\rm{2}}Baidu Inc. \\
    ycshensudaer@stu.suda.edu.cn, lingyongy@gmail.com, mcao@suda.edu.cn
}

\begin{document}

\maketitle

\begin{abstract}
Visual Retrieval-Augmented Generation (VRAG) empowers models to navigate and answer queries about visually rich documents by retrieving relevant page images as visual evidence and reasoning over their content.
However, effectively utilizing this visual evidence is usually impeded by two main challenges.
First, answer-relevant evidence is sparse and may be concentrated in a small region of one page or dispersed across multiple pages.
Second, existing agentic methods often generate answers based on raw exploration trajectories or compressed textual memories rather than an explicitly organized set of supporting images, making answers susceptible to exploration noise and obscuring the evidence-backed reasoning trace.
We argue that the bottleneck lies not only in evidence discovery but also in its preservation and organization before answer generation.
We propose \ourmethod{} (\textbf{S}election and \textbf{CO}nsolidation for \textbf{R}obust \textbf{E}vidence), a unified agent loop for explicit evidence selection and consolidation.
During exploration, \ourmethod retains only query-relevant observations and their source pointers in a maintained textual ledger, preserving earlier evidence while keeping the visual context bounded.
At termination, it reloads the referenced original images and consolidates the visual evidence for answering, arranging it into a logical sequence.
This decouples final reasoning from exploratory trial-and-error while ensuring strict visual grounding via indexed claim-to-image linkages.
To enable end-to-end optimization of this unified rollout, our training paradigm combines filtered cold-start trajectory distillation with evidence-aware reinforcement learning, whose reward promotes evidence coverage, consolidation compactness, and answer correctness.
Experiments on three established VRAG benchmarks demonstrate that \ourmethod achieves state-of-the-art overall accuracy on each benchmark across various backbone scales, with further gains from both training stages.
\end{abstract}


\section{Introduction}

Visual Retrieval-Augmented Generation (VRAG)~\cite{wang2026vrag,
wang2025vidorag, shen2026visor} extends traditional retrieval-augmented generation~\cite{arslan2024survey} to visually rich documents such as slides, reports, and scanned PDFs, retrieving and reasoning over page images. In these documents, evidential content is encoded not only in textual form but also through charts, tables, layout structures, and spatial relationships. Recent advances in Vision-Language Models (VLMs)~\cite{bai2025qwen3,
bai2025qwen2, liu2024deepseek, liu2023visual} enable VRAG methods like
VRAG-RL~\cite{wang2026vrag} to leverage raw visual inputs, preserving critical layout, structural, and multimodal cues that are often lost or distorted by traditional OCR-based methods~\cite{zhang2025ocr}.

\begin{figure}[t]
\centering
\includegraphics[width=\columnwidth]{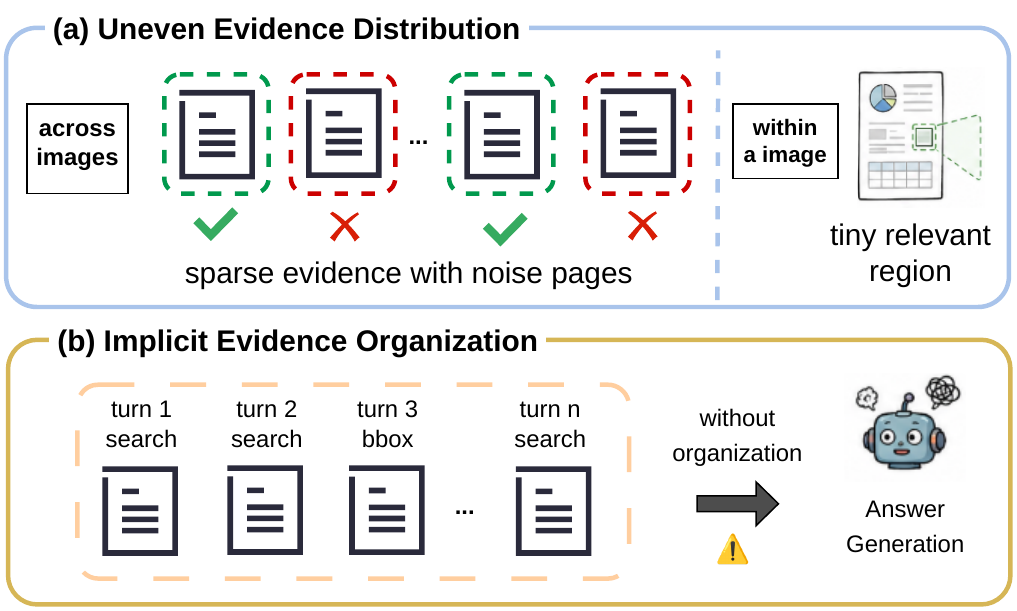}
\caption{Illustration of two challenges in 
visual evidence utilization in VRAG: (a) uneven evidence distribution; (b) implicit evidence organization.}
\label{fig:issue}
\end{figure}

In common VRAG settings, rendered document pages serve as the basic retrieval unit, resulting in a substantially coarser granularity than passage-level text retrieval. This page-level retrieval paradigm poses two practical challenges for effective visual evidence utilization.
(1) \texttt{Uneven evidence distribution.} Answer-relevant evidence may be localized within a small region of a single page or scattered across multiple different pages (Figure~\ref{fig:issue}~(a)). The agent must therefore determine which pages are relevant, how many of them are needed, and which regions require closer inspection.
(2) \texttt{Implicit evidence organization.} Evidence is often discovered incrementally across multiple retrieval and inspection steps. However, the discovered observations may remain organized according to the exploration process itself rather than the logical structure required for coherent answering (Figure~\ref{fig:issue}~(b)).

Existing agentic methods~\cite{wang2026vrag, shen2026visor} improve evidence discovery through iterative search and region-level zoom-in. VRAG-RL retains visual observations in the interaction trajectory, whereas VISOR distills them into a structured textual evidence space. However, neither explicitly reconstructs the original visual evidence into a compact, logically ordered chain for final answer generation or establishes claim-level links to visual sources. These limitations motivate our central insight: \textbf{exploration should discover potential evidence, while consolidation should select, restore, and organize it into an answer-oriented visual evidence chain for grounded generation.}

\begin{figure*}[htbp]
\centering
\includegraphics[width= 0.93\textwidth]{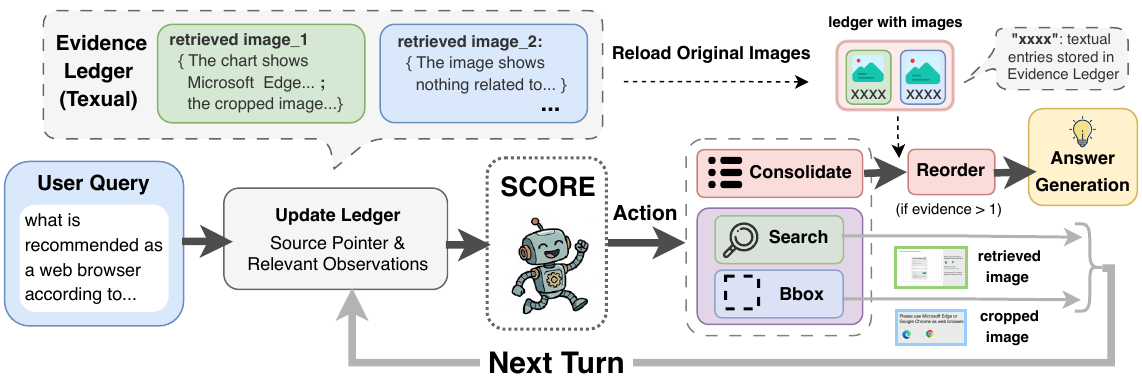}
\caption{Overview of \ourmethod{}. The agent iteratively searches page
images or zooms into local regions, updating a maintained textual
evidence ledger through observation summarization and relevance judgment.
Once the evidence is sufficient, the in-loop \texttt{consolidate} action
reloads the corresponding original images, filters and reorders the
retained evidence, and passes the organized visual context to final
answer generation within the same rollout.}

\label{fig:overview}
\end{figure*}

To address this bottleneck, we propose \ourmethod{} (\textbf{S}election and \textbf{CO}nsolidation for \textbf{R}obust \textbf{E}vidence), a unified agent loop for evidence organization in VRAG (Figure~\ref{fig:overview}).
The agent explores visual content through coarse image retrieval and fine-grained bounding-box zoom-in, treating full pages and cropped regions under a consistent relevance criterion. After each inspection, only query-relevant observations, along with precise source pointers, are recorded in a maintained textual ledger. This compact representation preserves essential evidence while deliberately bounding the accumulation of raw visual history.
When exploration terminates through a \texttt{consolidate} action or at
the turn limit, the referenced original page images are reloaded. 
The reloaded visual evidence is then consolidated and, when needed, arranged into a logical sequence for answering.
The final answer is generated from these images within the same
rollout, with indexed entries linking claims to supporting evidence.
It can be seen that \ourmethod separates exploratory trial and error from the final evidence chain, reducing exploration noise during answer generation while preserving visual grounding.

To further enforce structured evidence utilization, we introduce process-level supervision that explicitly shapes intermediate evidence organization rather than only final-answer correctness. 
During cold-start training, a teacher model generates trajectories containing relevance decisions, consolidated evidence order, and final answers; only those with full gold-page coverage and answers judged correct by an LLM evaluator are retained.
In reinforcement learning, an evidence-aware reward is proposed to jointly optimize gold-page coverage, compactness of the consolidated chain, and answer correctness. 
This two-stage training pipeline encourages the agent to preserve and organize relevant visual evidence throughout the entire rollout, thereby supporting reliable answer generation.
To assess the resulting model, we evaluate \ourmethod{} on three established VRAG benchmarks: ViDoSeek~\cite{wang2025vidorag},
SlideVQA~\cite{tanaka2023slidevqa}, and
MMLongBench~\cite{ma2024mmlongbench}.

Our main contributions are summarized as follows:

\begin{itemize}
    \item We propose \ourmethod{}, a unified agent that selects relevant visual evidence across pages and regions, then explicitly consolidates it by reloading, denoising, and ordering original images within the same rollout before answering.

    \item We introduce process-level supervision: cold-start from high-coverage teacher trajectories and an evidence-aware RL reward that jointly optimizes gold coverage, consolidation compactness, and answer correctness.

    \item Evaluated on three VRAG benchmarks, \ourmethod{} achieves state-of-the-art accuracy across both 3B and 7B backbones, while ablations confirm the complementary contributions of selection and consolidation, and further analyses demonstrate robust retrieval and favorable efficiency.

\end{itemize}
 
\section{Related Work}
\subsection{Visual Retrieval-Augmented Generation}

Retrieval-augmented generation (RAG) has shown strong effectiveness on knowledge-intensive tasks~\cite{lewis2020retrieval,gao2023retrieval,yang2024crag}, where traditional text-based approaches generate answers grounded in passages retrieved from textual corpora.
However, with the wide adoption of visually rich documents such as slides, reports, and scanned PDFs, knowledge is no longer confined to plain text. 
This motivates Visual RAG (VRAG), which extends RAG to retrieve and reason over visual document pages.
Early methods rely on OCR or document parsing to extract text from images, but such pipelines are lossy and fail to preserve layout, charts, and figures~\cite{zhang2025ocr}. 
More recently, OCR-free retrieval methods directly align text queries
with page images: ColPali~\cite{faysse2024colpali} introduces
late-interaction visual retrieval via token-level similarity between
query text and image patch embeddings, while
EVisRAG~\cite{sun2025visrag} feeds retrieved page images directly into a
VLM for multi-image understanding. However, constrained by top-$k$
retrieval, these methods often introduce irrelevant pages and cannot
adaptively refine retrieval based on prior observations, making it
difficult to focus on sparse query-relevant evidence.

\subsection{Agentic VRAG with Reinforcement Learning}

The agentic RAG paradigm was first introduced by ReAct~\cite{yao2022react}, which interleaves reasoning and action so that the model can decide when and how to retrieve on the fly. Building on the recent success of reinforcement learning for LLM reasoning~\cite{shao2024deepseekmath}, VRAG-RL~\cite{wang2026vrag} extends this idea to visual documents by defining a visual perception action space with cropping and zooming, and training the VLM with multi-turn RL to actively explore evidence across pages.
For multi-image evidence management, VISOR~\cite{shen2026visor} maintains a textual evidence ledger during agent interaction, recording query-relevant visual observations as text to suppress noise from irrelevant images over long horizons. LAT~\cite{liu2026look} instead targets evidence attribution, jointly optimizing reasoning and visual grounding within a single agent via RL, so that answers come with verifiable visual sources.
These methods improve visual exploration, memory, or attribution, but
do not explicitly reconstruct and globally organize a selected set of
original page images before answering. A complementary line of work
decomposes multi-image reasoning across specialized agents, e.g.,
ViDoRAG~\cite{wang2025vidorag} assigns planning, retrieval, and answering
to separate agents in an actor-critic loop. Such pipelines, however, are
difficult to optimize end-to-end and often incur additional orchestration
costs. \ourmethod{} instead makes evidence organization explicit within
a unified end-to-end agent loop.

\section{Method}
\subsection{Task Formulation}
Let $q$ denote a natural language query and
$\mathcal{C}=\{I_1,I_2,\ldots,I_N\}$ a large-scale corpus of $N$ page
images extracted from visually rich documents such as slides and reports.
The task is to produce an answer $a$ by iteratively searching
$\mathcal{C}$ and reasoning over the retrieved pages in the context of $q$. Crucially,
$\mathcal{C}$ forms a flat, document-agnostic image pool in which pages
from heterogeneous documents are intermixed without document-level scoping.
Relevant evidence answering $q$ may reside entirely within a single page or be distributed across multiple pages,
while pages from unrelated documents serve as distractors; the model
must retrieve and aggregate the relevant visual signals when necessary.

\subsection{\ourmethod Framework}\label{sec:overview}
As illustrated in Figure~\ref{fig:overview} and
Algorithm~\ref{alg:SCORE}, \ourmethod operates as a unified agent loop with two persistent state objects: a maintained \textbf{textual evidence ledger} $\mathcal{L}$ and an interaction history $\mathcal{H}$. At each reasoning turn, the agent processes the current page or cropped region and produces a structured output comprising three fields: $\langle$\texttt{observe}$\rangle$ (a summary of the visual content), $\langle$\texttt{relevant}$\rangle$ (a binary relevance judgment), and $\langle$\texttt{action}$\rangle$ (the next operation, detailed below). Relevant observations are added to $\mathcal{L}$ along with pointers to their corresponding source images, while only the most recent $W$ raw visual observations from $\mathcal{H}$ remain in the active context. When exploration ends, the original images referenced in $\mathcal{L}$ are reloaded and organized for final answering. Keeping exploration, evidence maintenance, organization, and answering inside one loop forms a continuous rollout optimizable with a trajectory-level objective.

\begin{algorithm}[t]
\caption{\ourmethod Agent Loop}\label{alg:SCORE}
\begin{algorithmic}[1]
\REQUIRE Query $q$, corpus $\mathcal{C}$, max turns $T$, window $W$
\STATE $\mathcal{L}\leftarrow\emptyset$, $\mathcal{H}\leftarrow[\,]$,
       $(\operatorname{src}_1,o_1)\leftarrow(\varnothing,\varnothing)$
\FOR{$t=1$ \TO $T$}
    \STATE Context $\leftarrow$ prompt$(q)$, $\mathcal{L}$, current
           observation $o_t$, and the last $W-1$ turns of $\mathcal{H}$
    \STATE Generate $(\tilde{o}_t,\rho_t,a_t)$; at $t=1$, use
    $(\texttt{no image},\texttt{no},\texttt{search}(q))$
    \IF{$\rho_t=\texttt{yes}$}
        \STATE Add $(\operatorname{src}_t,\tilde{o}_t)$ to $\mathcal{L}$
    \ENDIF
    \STATE Set $a_t\leftarrow\texttt{consolidate}$ if $t=T$
    \IF{$a_t=\texttt{search}$}
        \STATE $(\operatorname{src}_{t+1},o_{t+1})\leftarrow$
        top-1 page retrieved from $\mathcal{C}$
    \ELSIF{$a_t=\texttt{bbox}$}
        \STATE $(\operatorname{src}_{t+1},o_{t+1})\leftarrow$
        bbox cropped region 
    \ELSIF{$a_t=\texttt{consolidate}$}
        \STATE $\mathcal{E}\leftarrow$ original images referenced by $\mathcal{L}$
        \STATE If $|\mathcal{E}|>1$, select, denoise, and order $\mathcal{E}$
        \STATE Generate \texttt{answer} action over visual evidence $\mathcal{E}$
        \RETURN final answer $a$
    \ENDIF
    \STATE Append $(o_t,\tilde{o}_t,\rho_t,a_t)$ to $\mathcal{H}$
\ENDFOR
\end{algorithmic}
\end{algorithm}

\paragraph{Context Construction.}
Following VISOR~\cite{shen2026visor}, we control the growth of raw visual context by reconstructing the input at each turn $t$ from a fixed prompt, the evolving evidence ledger, and a bounded window of recent interactions. Let $\mathbf{C}_t$ denote the active context supplied to the agent at turn $t$; it is formed by concatenating these components as follows:
\begin{equation}\label{eq:context}
\mathbf{C}_t=
\bigl[\,\mathbf{P}_{\text{init}};\mathcal{L}_t;
\mathcal{H}_{t-W+1:t-1};o_t\,\bigr],
\end{equation}
where $\mathbf{P}_{\text{init}}$ contains the system prompt and user query $q$, $\mathcal{L}_t$ is the ledger of relevance-filtered summaries paired with source image pointers, $\mathcal{H}_{t-W+1:t-1}$ contains the most recent $W-1$ completed turns, and $o_t$ is the current page or cropped region under consideration. 
We use $W=2$ to preserve
the latest retrieve-then-zoom chain. Raw visual inputs beyond the sliding window are evicted to bound context usage, while semantically distilled evidence persists in $\mathcal{L}_t$. 
The context includes an intent-injection reminder that restates $q$ and points to the collected evidence. Details for the prompt and intent-injection reminder are provided in
the \textbf{Appendix}.

\paragraph{Action space.}
The agent selects from three structured actions at each step:
\textbf{(1)} $\langle$\texttt{search}$\rangle$ issues a textual query to a vision-aware retriever and returns the top-ranked page from $\mathcal{C}$ as the next \emph{retrieved image}. The initial search is constrained to the original user query $q$ to prevent premature query reformulation before sufficient evidence is gathered, and subsequent searches may generate refined sub-queries conditioned on the evolving ledger $\mathcal{L}_t$;
\textbf{(2)} $\langle$\texttt{bbox}$\rangle$ specifies a bounding box on the current page to obtain a cropped image, enabling fine-grained inspection of visual regions;
\textbf{(3)} $\langle$\texttt{consolidate}$\rangle$ terminates exploration and triggers answer generation. It reloads the original full-page images referenced in $\mathcal{L}_t$, reorders and filters them if multiple pages are retained, and forwards the curated evidence to the final-answer module, which produces the terminal  ⟨answer⟩ action containing the model’s response.

\subsection{Evidence Ledger: Selection and Consolidation}

\ourmethod organizes sparse evidence in two stages. During exploration,
relevant observations, along with pointers to their source images, are incrementally accumulated in the compact textual ledger; at termination, the \texttt{consolidate} action reloads the original images referenced in the ledger and globally reorganizes them to support holistic visual reasoning during answer generation. This design ensures that the ledger acts as a bridge between bounded, sequential exploration and the final, evidence-grounded visual-semantic generation.

\paragraph{Source-linked visual evidence selection.}
At exploration turn $t$, the agent outputs a binary relevance label
$\rho_t\in\{\texttt{yes},\texttt{no}\}$ along with a textual summary $\tilde{o}_t$  of  the current visual observation in \texttt{<observe>}. 
The ledger is updated by
\begin{equation}
\mathcal{L}_t =
\begin{cases}
\mathcal{L}_{t-1}\mathbin{\|}
\bigl[(\operatorname{src}_t,\tilde{o}_t)\bigr],
& \rho_t=\texttt{yes},\\
\mathcal{L}_{t-1}, & \rho_t=\texttt{no},
\end{cases}
\end{equation}
where $\mathbin{\|}$ appends a new entry to the ledger, and
$\operatorname{src}_t$ identifies the original page and, for a cropped observation, its bbox coordinates. Each retained entry thus links the textual observation $\tilde{o}_t$ to a visual source that can be reloaded during consolidation. Observations marked \texttt{no} are not added to $\mathcal{L}$; they remain only temporarily in the bounded visual context and are removed as the window advances.

\paragraph{Global evidence consolidation.}
Turn-wise relevance judgments in Eq. (2) may still yield redundant or suboptimally ordered ledger entries. 
Upon triggering $\langle$\texttt{consolidate}$\rangle$ or reaching the maximum turn limit $T$, the agent reloads the original images referenced in the terminal ledger $\mathcal{L}_{\tau}$, where
$\tau$ denotes the final exploration step. 
For multiple images, the agent examines the reloaded images together with respect to $q$, retains the useful ledger entries, and arranges them in a logical order for answering.
When only one image is involved, no further selection or reordering is needed, and the terminal ledger is used directly.
The agent then answers $q$ using the visual evidence associated with the resulting ledger entries.
In its final output, the agent indicates which ledger entry supports each claim, thereby linking the answer to the selected visual evidence rather than the noisy exploration trajectory.

\subsection{Training Pipeline}\label{sec:training}

Since per-turn relevance judgment and global consolidation are not directly supervised by answer correctness, we adopt a two-phase training pipeline: a \emph{cold-start} phase that instills the structured output format and basic ledger behavior, followed by a \emph{reinforcement learning} phase whose reward explicitly targets evidence selection alongside final-answer quality.

\paragraph{Cold-Start via Filtered Trajectory Distillation.} We distill complete \ourmethod trajectories generated by a stronger teacher, Qwen3.5-122B-A10B~\cite{qwen3.5}, on training queries from SlideVQA~\cite{tanaka2023slidevqa}. We retain only trajectories with both LLM-judged correct answers and full evidence coverage, requiring every gold reference page to appear in the final answer ledger rather than merely being retrieved. This filter removes superficially correct trajectories that happen to omit essential supporting evidence. We then perform SFT with assistant-only label masking to teach the structured output format and initialize ledger management and consolidation. Further details on rollout, filtering, and SFT conversion are provided in the \textbf{Appendix}.

\begin{table*}[htbp]
\centering
\resizebox{\textwidth}{!}{%
\begin{tabular}{lccc|ccc|cccccc}
\toprule
\multirow{2}{*}{Method} &
  \multicolumn{3}{c}{SlideVQA} &
  \multicolumn{3}{c}{ViDoSeek} &
  \multicolumn{6}{c}{MMLongBench} \\
\cmidrule(lr){2-4}\cmidrule(lr){5-7}\cmidrule(lr){8-13}
 & Single-hop & Multi-hop & Overall & Extraction & Logic & Overall & Text & Table & Chart & Figure & Layout & Overall \\
\midrule
\rowcolor{gray!15}
\multicolumn{13}{c}{\textit{Qwen2.5-VL-7B}} \\
\midrule
Vanilla RAG~\cite{faysse2024colpali}      & 29.10 & 17.40 & 26.10 & 26.40 & 41.30 & 32.88 & 13.10 & 14.70 & 15.90 & 4.30  & 7.60  & --    \\
ReAct~\cite{yao2022react}                 & 34.80 & 20.40 & 31.11 & 27.50 & 42.10 & 33.85 & 10.10 & 12.40 & 10.20 & 6.20  & 7.10  & --    \\
ViDoRAG$^{\dagger\star}$~\cite{wang2025vidorag}          & 72.15 & 39.86 & 63.88 & 66.05 & 72.83 & 69.00 & 24.40 & 23.96 & 21.91 & 24.14 & 20.34 & 25.50 \\
M3RAG$^\dagger$~\cite{du2026m3rag}          & --    & --    & 65.82 & --    & --    & 69.36 & --    & --    & --    & --    & --    & --    \\
Search-R1-VL$^\ddagger$~\cite{jin2025search}  & 48.30 & 42.30 & 46.76 & 40.50 & 50.30 & 44.77 & 19.90 & 13.40 & 12.90 & 11.40 & 10.20 & --    \\
VRAG-RL$^\ddagger$~\cite{wang2026vrag}        & 69.30 & 43.10 & 62.59 & 60.60 & 74.80 & 66.78 & 26.10 & \underline{26.30} & 24.80 & 25.90 & 21.20 & --    \\
MMSearch-R1$^{\ddagger\star}$~\cite{wu2025mmsearch}          & 52.06 & 40.21 & 49.03 & 55.97 & 59.56 & 57.53 & 16.84 & 17.97 & 19.10 & 18.28 & 11.86 & 18.42 \\
EVisRAG$^{\ddagger\star}$~\cite{sun2025visrag}               & 78.21 & 42.32 & 69.09 & 67.75 & 72.43 & 69.79 & 26.80 & \textbf{27.65} & 24.72 & 22.41 & 19.49 & 27.98 \\
R1-Router$^{\ddagger\star}$~\cite{peng2025learning}   & 69.66 & 45.33 & 63.43 & 64.19 & 68.61 & 66.11 & 26.46 & 23.50 & 22.47 & \underline{28.28} & 16.95 & 26.92 \\
VISOR$^{\ddagger}$~\cite{shen2026visor}                          & \underline{78.82}  & \underline{53.62} & \underline{72.37}  & \textbf{73.49}  & \underline{76.66} & \underline{74.87}  & \underline{27.49} & 23.96  & \underline{27.53}  & 23.79  & \underline{22.88}  & \underline{28.45}  \\
\ourmethod (ours)$^\ddagger$                          & \textbf{82.28}  & \textbf{62.26} & \textbf{77.16}  & \underline{73.33}	& \textbf{77.87} & \textbf{75.31}  & \textbf{32.30}	& 25.35	& \textbf{31.46}	& \textbf{30.00}	& \textbf{27.97}  & \textbf{31.52}  \\
\midrule
\rowcolor{gray!15}
\multicolumn{13}{c}{\textit{Qwen2.5-VL-3B}} \\
\midrule
Vanilla RAG~\cite{faysse2024colpali}      & 19.40 & 12.20 & 17.56 & 10.10 & 17.30 & 13.23 & 2.20  & 4.10  & 5.20  & 4.70  & 4.30  & --    \\
ReAct~\cite{yao2022react}                 & 15.70 & 10.90 & 14.47 & 6.70  & 14.20 & 9.96  & 2.70  & 3.60  & 3.40  & 3.10  & 5.10  & --    \\
ViDoRAG$^{\dagger\star}$~\cite{wang2025vidorag}          & 41.44 & 19.93 & 35.94 & 31.32 & 38.23 & 34.33 & 7.90  & 6.91  & 5.06  & 8.97  & 7.63  & 8.50  \\
Search-R1-VL$^\ddagger$~\cite{jin2025search}  & 26.30 & 20.10 & 24.71 & 20.10 & 29.80 & 24.32 & 8.50  & 7.80  & 7.90  & 9.30  & 7.60  & --    \\
VRAG-RL$^\ddagger$~\cite{wang2026vrag}        & 65.30 & 38.60 & 58.45 & 63.10 & \textbf{73.80} & 67.76 & 22.70 & 16.10 & 21.90 & 21.40 & 19.50 & --    \\
EVisRAG$^{\ddagger\star}$~\cite{sun2025visrag}               & \underline{75.42} & 47.70 & 68.35 & 66.05 & \underline{72.43} & 68.82 & \underline{27.14} & \textbf{28.64} & 25.13 & \underline{24.83} & 17.80 & \underline{28.34} \\
R1-Router$^{\ddagger\star}$~\cite{peng2025learning}   & 64.93 & 42.15 & 59.10 & 62.64 & 69.42 & 65.59 & 26.80 & 23.50 & 21.91 & \textbf{25.17} & \underline{21.19} & 25.74 \\
VISOR$^{\ddagger}$~\cite{shen2026visor}                                 & 74.58  & \underline{50.79} & \underline{68.49}  & \textbf{67.75}  & 70.62 & \underline{69.00}  & 26.80  & \underline{23.96}  & \underline{26.97}  & 23.45  & \textbf{22.03}  & 27.86  \\
\ourmethod (ours)$^\ddagger$                   & \textbf{76.33}  & \textbf{53.44} & \textbf{70.47}  & \underline{67.29} & 71.83 & \textbf{69.26}  & \textbf{27.49} & \underline{23.96} & \textbf{27.53} & 23.79 & \underline{21.19}  & \textbf{28.51}  \\
\bottomrule
\end{tabular}%
}
\caption{Main results on SlideVQA, ViDoSeek, and MMLongBench. We report accuracy (\%). $\dagger$ denotes multi-agent architectures. $\ddagger$ denotes fine-tuned models. $\star$ denotes results reproduced under our experimental settings for a fair comparison; all other results are cited from the original papers. The best result in each column is \textbf{bolded} and the second-best is \underline{underlined}.}
\label{tab:main}
\end{table*}

\paragraph{Reinforcement Learning.}
Building upon the cold-start checkpoint, we perform multi-turn RL using GRPO~\cite{shao2024deepseekmath}. 
Since \ourmethod realizes exploration, consolidation, and answering within a single continuous rollout under a unified sliding-window context, a single trajectory-level optimization suffices to jointly train these capabilities. This process is driven by a novel \textbf{evidence-aware trajectory reward}, which simultaneously evaluates evidence selection quality and final answer correctness.
Formally, let $\mathcal{P}(\mathcal{L})=\{\,\text{page}(\text{src}):
(\text{src},\tilde{o})\in\mathcal{L}\,\}$ denote the source pages backing a ledger $\mathcal{L}$. 
We define $\mathcal{P}^{\star}$ as the set of gold reference pages and $\mathcal{P}_{\text{ans}}=\mathcal{P}(\mathcal{L}_{\text{ans}})$ as the pages selected in the final answer ledger. 
We introduce two metrics:
\begin{equation}
\textit{cov} = \frac{|\mathcal{P}^{\star}\cap\mathcal{P}_{\text{ans}}|}{|\mathcal{P}^{\star}|},\quad
\textit{cmp} = \frac{|\mathcal{P}^{\star}\cap\mathcal{P}_{\text{ans}}|}{|\mathcal{P}_{\text{ans}}|},
\end{equation}
where \textit{cov} (Recall) rewards the retention of gold pages and \textit{cmp}
(Precision) rewards a compact ledger by penalizing irrelevant retrievals.
We define \textit{cmp}=0 if $\mathcal{P}_{\text{ans}}$ is empty. 
The trajectory reward combines evidence selection with answer correctness:
\begin{equation}\label{eq:reward}
r = \textit{cov} + \lambda_{\text{cmp}}\cdot\mathbf{I}_{\text{cov}=1}\cdot\textit{cmp} + \mathbf{I}_{\text{cov}=1}\cdot r_{\text{ans}} - \mathbf{I}_{\text{cov}<1}\cdot\beta,
\end{equation}
where $\lambda_{\text{cmp}}{=}0.2$ and $\beta{=}1$ are hyperparameters,
and the indicator $\mathbf{I}_{\text{cov}=1}$ equals $1$ when the ledger
covers all gold pages and $0$ otherwise ($\mathbf{I}_{\text{cov}<1}$ is
its complement). Through this indicator, both the compactness term and the
answer reward $r_{\text{ans}}$ (a binary LLM-as-judge label) are gated on full
coverage, while incomplete coverage instead incurs the penalty $\beta$. This
gating prevents the agent from gaming \textit{cmp} by dropping evidence or
receiving answer credit for an incomplete ledger. Consequently, this evidence-aware signal aligns with \ourmethod's core objective: compelling the agent to recover all gold pages while filtering distractors, even under sparse and noisy visual conditions.
During RL, retrieved visual observations and system-injected tokens are masked from the policy loss. The agent decodes from the sliding-window context, while GRPO aggregates the loss over all agent-generated tokens under their corresponding rolling contexts using the shared trajectory-level reward.

\section{Experiments}
\subsection{Experimental Settings}

\paragraph{\textbf{Datasets and Metric.}} 
We evaluate \ourmethod on three established benchmarks for VRAG: \textbf{ViDoSeek}~\cite{wang2025vidorag}, \textbf{SlideVQA}~\cite{tanaka2023slidevqa}, and \textbf{MMLongBench}~\cite{ma2024mmlongbench}. We follow the unified-corpus protocol of VRAG-RL~\cite{wang2026vrag}: all document pages are flattened into a single image pool, and the model must retrieve the relevant pages from this shared corpus to answer each query, without any document-level scoping at inference time. Training data is drawn from the SlideVQA training split and consists of 2{,}500 trajectories used for cold-start distillation and 1{,}600 queries used for reinforcement learning. We adopt the same LLM-judge evaluation as prior work~\cite{wang2026vrag}: \texttt{Qwen-max-latest}~\cite{qwen3.5} compares each predicted answer with the reference and outputs a binary correctness score, and we report the mean as accuracy. More details are in the \textbf{Appendix}.

\paragraph{\textbf{Baselines.}} 
We benchmark \ourmethod against two method families: (1) \emph{vanilla} (non-fine-tuned) approaches, including Vanilla RAG~\cite{faysse2024colpali}, ReAct~\cite{yao2022react}, and multi-agent pipelines ViDoRAG~\cite{wang2025vidorag} and M3RAG~\cite{du2026m3rag}; and (2) \emph{fine-tuned} methods using task-specific supervision like \ourmethod: Search-R1-VL~\cite{jin2025search}, VRAG-RL~\cite{wang2026vrag}, MMSearch-R1~\cite{wu2025mmsearch}, R1-Router~\cite{peng2025learning}, and EVisRAG~\cite{sun2025visrag}. To isolate agent design contributions, all reproduced methods use the ColQwen2.5-v0.1~\cite{faysse2024colpali} retrieval backbone, while adopted results retain their original settings (compatibility in the \textbf{Appendix}).

\subsection{Main Results}

As shown in Table~\ref{tab:main}, \ourmethod achieves the best overall accuracy across all three benchmarks and both backbone sizes.
Using Qwen2.5-VL-7B as the backbone, it improves over the strongest prior baseline from 72.37\% to 77.16\% on SlideVQA, 74.87\% to 75.31\% on ViDoSeek, and 28.45\% to 31.52\% on MMLongBench; the same trend holds with the smaller Qwen2.5-VL-3B.
Unlike prior methods that reason directly over retrieved pages or search trajectories, \ourmethod explicitly selects and organizes sparse evidence before answering, leading to consistent gains across benchmarks.

The gains are largest when evidence must be aggregated across pages or localized within complex layouts.
On SlideVQA, \ourmethod improves most on multi-hop questions (62.26\% vs.\ 53.62\% from VISOR under the 7B backbone), while its single-hop gain is more moderate.
On MMLongBench, where answers rely on sparse evidence scattered across pages and confined to small regions, \ourmethod achieves the best 7B results on Text, Chart, Figure, and Layout. This stems from: (1) selective ledger updates that choose which pages to retain, and (2) bbox zoom-ins that select which regions to read—jointly addressing the uneven spatial and document-level distribution of evidence.
Overall, these results demonstrate the effectiveness of explicitly organizing sparse visual evidence beyond merely retrieving it in VRAG.

\subsection{Ablation Study}

We ablate the two evidence-organization modules in \ourmethod: relevance
judgment, which filters observations at collection time, and
consolidation, which globally re-selects and reorders the ledger
before answering. The variant \emph{w/o both} removes both, degrading \ourmethod into a
plain accumulate-all sliding-window agent. We report each variant under
both the untrained \emph{Vanilla} setting and the \emph{Fine-tuned}
(cold-start\,+\,RL) setting, on ViDoSeek and SlideVQA with the
Qwen2.5-VL-7B backbone under the main-result protocol.

\begin{table}[!h]
\centering
\resizebox{\columnwidth}{!}{%
\begin{tabular}{lcccc}
\toprule
\multirow{2}{*}{Variant} &
  \multicolumn{2}{c}{SlideVQA} & \multicolumn{2}{c}{ViDoSeek} \\
\cmidrule(lr){2-3}\cmidrule(lr){4-5}
 & Vanilla & Fine-tuned & Vanilla & Fine-tuned \\
\midrule
\ourmethod (Full)   & \textbf{55.17} & \textbf{77.16} & \textbf{48.25} & \textbf{75.31} \\
\quad w/o relevance judgment   & 53.50 & 72.28 & 47.72 & 71.80 \\
\quad w/o consolidation & 45.24 & 74.13 & 36.69 & 73.91 \\
\quad w/o both & 44.06 & 71.51 & 35.73 & 69.79 \\
\bottomrule
\end{tabular}%
}
\caption{Ablation of the two organization modules. Accuracy (\%) on the Qwen2.5-VL-7B under the Vanilla and Fine-tuned (cold-start\,+\, RL) settings.
Best per column in \textbf{bold}.}
\label{tab:ablation}
\end{table}

Ablating either module hurts performance, and removing \emph{both} is consistently worst, confirming their complementarity. Their relative importance \emph{flips} after fine-tuning. In the
\emph{Vanilla} setting, dropping consolidation is far more damaging than
dropping relevance judgment (e.g.\ $-9.9$ vs.\ $-1.7$ on SlideVQA): without training, the model judges relevance poorly during the intermediate turns, so substantial noise enters the ledger, making global consolidation the more critical safeguard for filtering it before answer generation.
After \emph{Fine-tuning},
the trend reverses: dropping relevance judgment becomes the larger drop
(e.g.\ $-4.9$ vs.\ $-3.0$ on SlideVQA). 
Since the model has learned to filter evidence at collection time, keeping the ledger clean. Without this early filtering, consolidation must instead process raw, unfiltered retrievals.
In effect, training shifts the
denoising responsibility from the exit (consolidation) to the entrance
(relevance judgment), yet both modules remain beneficial throughout.
The necessity of other components of \ourmethod, including the maintained ledger (vs. a sliding window), window size W, intent injection, and bbox zoom-in, is further analyzed in the \textbf{Appendix}.

\subsection{More Analysis}

\paragraph{Retrieval and Consolidation Behavior Analysis.}
We analyze \emph{where} accuracy comes from using the retrieval and evidence-organization metrics reported in Table~\ref{tab:retrieval_completeness}.
Completeness measures the percentage of queries whose full retrieval trajectory contains every gold page, whereas ledger coverage measures the percentage whose final ledger retains every gold page.
Two observations stand out.
\ding{182} Achieving high completeness incurs prohibitive costs without ensuring superior performance. 
ViDoRAG achieves the highest completeness (91.3) only by retrieving 10 fixed images, which introduces many non-gold pages into the visual context. In contrast, VRAG-RL reduces non-gold evidence but misses relevant pages. This reveals a clear \textit{coverage--noise tension}: models struggle to retrieve all relevant pages without including noise.
\ding{183} Explicit evidence organization better resolves this tension. VISOR reduces visual noise by converting pages into a textual ledger, but this conversion may introduce transcription errors and, without further selection, still retains $1.29$ non-gold pages per query. In contrast, \ourmethod keeps only selected page \emph{images}, avoiding conversion loss while cutting non-gold pages to $0.16$, about one-fifth of the lowest image-based baseline. Despite a marginal drop in ledger coverage, this rigorous filtering yields the highest accuracy ($77.16$).
Together, these results show that selecting and organizing evidence, rather than merely retrieving more pages, is critical to accurate answering.

\begin{table}[!h]
\centering
\resizebox{\columnwidth}{!}{%
\begin{tabular}{lccccc}
\toprule
\textbf{Method} & \textbf{Comp.} & \textbf{Avg.\ Ret.} &
  \textbf{Ledger Cov.} & \textbf{Noise Img.} & \textbf{Acc.} \\
\midrule
EVisRAG~\cite{sun2025visrag} & 80.2 & 3 (fixed)  & -- & 1.97 & 69.09 \\
VRAG-RL~\cite{wang2026vrag}  & 76.8 & 1.78       & -- & 0.77 & 62.59 \\
ViDoRAG~\cite{wang2025vidorag} & \textbf{91.3} & 10 (fixed) & -- & 8.81 & 63.88 \\
VISOR~\cite{shen2026visor}   & 84.2 & 2.34       & \textbf{84.2} & 1.29 & 72.37 \\
\textbf{\ourmethod} & 81.9 & \textbf{1.76} & 78.4 & \textbf{0.16} & \textbf{77.16} \\
\bottomrule
\end{tabular}%
}
\caption{Retrieval and Consolidation Behavior Analysis on SlideVQA.
\textbf{Comp.}: retrieval completeness (\%); \textbf{Avg.\ Ret.}: average pages retrieved; \textbf{Ledger Cov.}: gold-page coverage of final evidence ledger (\%), where methods without an evidence ledger are marked ``--''; \textbf{Noise Img.}: average non-gold pages retained for answering; \textbf{Acc.}: answer accuracy (\%).}
\label{tab:retrieval_completeness}
\end{table}

\paragraph{Time Efficiency.}
We investigate whether the explicit consolidation and answering steps in \ourmethod incur a prohibitive computational cost. 
We have applied two mechanisms to constrain this overhead. 
First, we bound the raw visual context to the last $W{=}2$ turns, while retaining earlier relevant observations solely in the textual ledger. Since image tokens are significantly more expensive than text, this mechanism prevents visual token consumption from scaling with trajectory length during exploration while preserving essential evidence. 
Second, our learned policy converges in an average of $3.65$ turns, fewer than both ViDoRAG ($6.22$) and VRAG-RL ($4.36$), partially offsetting the consolidation cost.

\begin{table}[!t]
\centering
\resizebox{\columnwidth}{!}{%
\begin{tabular}{lcccc}
\toprule
\textbf{Method} & \textbf{Avg.\ Turns} & \textbf{Avg.\ Tokens} & \textbf{Latency (s)} & \textbf{Acc.}\\
\midrule
ViDoRAG~\cite{wang2025vidorag} & 6.22 & 23259 & 18.85 & 59\\
VRAG-RL~\cite{wang2026vrag}    & 4.36 & 2932 & 5.23 & 61\\
EVisRAG~\cite{sun2025visrag}   & \textbf{1} & \textbf{2514} & \textbf{4.37} & 61\\
VISOR~\cite{shen2026visor}     & 3.40 & 3162 & 5.68 & 66\\
\textbf{\ourmethod-direct}          & \underline{2.83} & \underline{2762} & \underline{4.97} & \underline{70}\\
\textbf{\ourmethod}                 & 3.65 & 3468 & 6.42 & \textbf{72}\\
\bottomrule
\end{tabular}%
}
\caption{Efficiency on SlideVQA (backbone: Qwen2.5-VL-7B): average agent turns (Avg. Turns),
tokens consumed (Avg. Tokens), and per-query latency (Latency),
averaged over 100 balanced cases (50 single-hop, 50 multi-hop).}
\label{tab:efficiency}
\end{table}

\begin{figure}[!t]
\centering
\includegraphics[width=0.7\columnwidth]{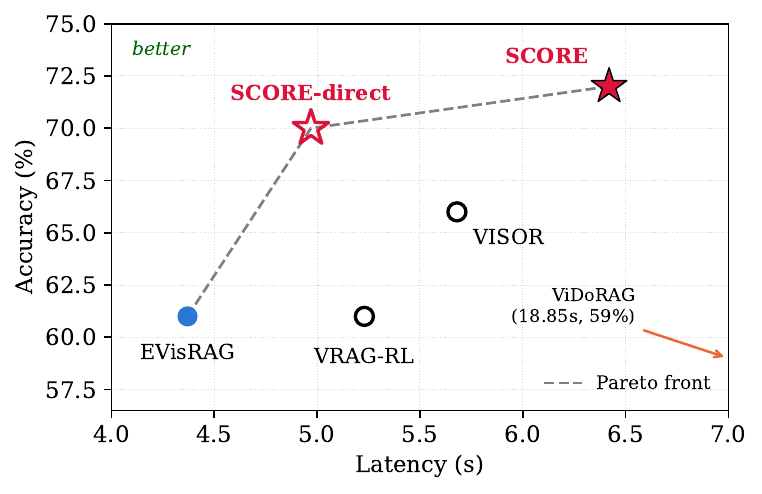}
\caption{Accuracy versus average latency per query on SlideVQA. Both
\ourmethod variants lie on the Pareto front, improving accuracy along
the frontier as latency increases.}
\label{fig:efficiency_pareto}
\end{figure}

Table~\ref{tab:efficiency} details the average agent turns, token consumption, and end-to-end latency per query, showing a favorable accuracy–latency trade-off. 
\textbf{\ourmethod-direct}, an efficient variant of \ourmethod that answers directly from the textual ledger without visual consolidation, achieves an accuracy of $70$ at $4.97$s. This outperforms all baselines in accuracy while remaining faster than all except EVisRAG. Compared to VISOR, it gains $4$ points while reducing latency by $0.71$s, suggesting that effective evidence organization contributes significantly beyond mere computational overhead. 
Incorporating consolidation and answering over reloaded original images (\ourmethod) further raises accuracy to $72$, with average increases of $1.45$s in latency and $706$ tokens. Notably, against the multi-agent ViDoRAG, \ourmethod uses $6.7\times$ fewer tokens while scoring $13$ points higher. Consequently, both variants reside on the Pareto front in Figure~\ref{fig:efficiency_pareto}.

\begin{table}[!b]
\centering
\resizebox{\columnwidth}{!}{%
\begin{tabular}{lcc|ccc}
\toprule
\multirow{2}{*}{\textbf{Variant}} & \multirow{2}{*}{\textbf{Comp.}} &
  \multirow{2}{*}{\textbf{Ledger Cov.}} & \multicolumn{3}{c}{\textbf{SlideVQA Acc.}} \\
\cmidrule(lr){4-6}
 & & & Single & Multi & Overall \\
\midrule
Base (prompting)        & 72.6 & 64.0 & 61.65 & 36.33 & 55.17 \\
\ + Cold-start (SFT)    & 80.8 & 76.6 & 80.46 & 61.55 & 75.62 \\
\ + Cold-start + RL     & \textbf{81.9} & \textbf{78.4} & \textbf{82.28} & \textbf{62.26} & \textbf{77.16} \\
\bottomrule
\end{tabular}%
}
\caption{Training-stage decomposition on SlideVQA (Qwen2.5-VL-7B).
Comp.: retrieval completeness; Ledger Cov.: gold pages kept in the
ledger (\%). Acc.\ is overall accuracy with single-/multi-hop splits.
All in \%. \emph{Base}: \ourmethod framework by prompting, no training.}
\label{tab:training}
\end{table}

\paragraph{Effect of Training.}
To disentangle the contributions of each training stage, we evaluate three variants of \ourmethod on SlideVQA (Table~\ref{tab:training}): 
(i)~an untrained \emph{Base} that executes the framework via prompting alone, 
(ii)~a \emph{Cold-start} checkpoint obtained through supervised fine-tuning, 
and (iii)~the full \emph{Cold-start+RL} model. 
In addition to accuracy, we report retrieval completeness (Comp.) and ledger coverage (Ledger Cov.) to diagnose \emph{what} each stage improves.
The dominant gain originates from cold-start SFT, which lifts overall accuracy from 55.17 to 75.62; RL contributes a further, more modest increment to 77.16. 
We attribute this to the construction of the cold-start data: each filtered trajectory encodes an explicit evidence--answer correspondence, in which every retained turn records a relevance decision and the consolidation step specifies which pages support the final answer. Because the filtering criterion admits only trajectories that yield correct answers while covering all gold pages, SFT can internalize this alignment directly. 
Consistent with this view, ledger coverage exhibits its sharpest increase at the SFT stage ($64.0\rightarrow76.6$), and multi-hop accuracy improves by 25.2 points ($36.3\rightarrow61.6$), confirming that questions spanning multiple pages benefit from keeping faithful evidence. 
RL, by contrast, refines evidence selection at the margin: coverage continues to rise ($\rightarrow78.4$) while completeness remains largely static, suggesting that RL sharpens an already-acquired retrieval policy rather than inducing the core behavior from scratch.

\begin{table}[!h]
\centering
\resizebox{\columnwidth}{!}{%
\begin{tabular}{lccc}
\toprule
\textbf{Backbone} & \textbf{VRAG-RL} & \textbf{\ourmethod} & \textbf{$\Delta$} \\
\midrule
Qwen2.5-VL-3B~\cite{bai2025qwen2}       & 16.79 & 28.58 & $+11.79$ \\
Qwen2.5-VL-7B~\cite{bai2025qwen2}       & 36.98 & 55.17 & $+18.19$ \\
Qwen3-VL-8B~\cite{bai2025qwen3}         & 63.97 & 70.84 & $+6.87$ \\
Qwen3.5-27B~\cite{qwen3.5}         & 81.85 & 82.53 & $+0.68$ \\
Qwen3.5-122B-A10B~\cite{qwen3.5}   & 80.72 & 82.93 & $+2.21$ \\
\bottomrule
\end{tabular}%
}
\caption{Effect of backbone strength on SlideVQA under a
\emph{prompting-only} setting (no training).The trained \ourmethod result is reported in
Table~\ref{tab:main}.}
\label{tab:backbone}
\end{table}

\paragraph{Backbone Analysis.}
We investigate how backbone capability influences the benefit of \ourmethod's explicit evidence organization. 
To isolate this effect, we compare against VRAG-RL~\cite{wang2026vrag}, a representative method that accumulates a retrieval trajectory by iteratively retrieving, zooming in, and answering from an ever-growing context, across five backbones spanning two model families (Qwen2.5-VL and Qwen3/3.5) and scales from 3B to 122B parameters, all under a prompting-only setting (no task-specific training). 
Table~\ref{tab:backbone} reports SlideVQA accuracy and the per-backbone gain $\Delta$.

\ourmethod improves every backbone, with $\Delta$ peaking at 7B ($+18.19$) and narrowing on stronger models, suggesting that they increasingly handle multi-image reasoning and noise internally.
The 3B model still gains
$11.79$ points but appears less able to execute consolidation effectively:
under prompting only, a 3B variant that keeps the textual ledger but drops
selection and consolidation even scores \emph{higher} ($42.26$) than full
\ourmethod ($28.58$), as the weaker backbone cannot yet apply these steps
reliably and instead mis-selects or mis-orders evidence.
The benefit is therefore largest at an intermediate capacity, where the
backbone can follow the explicit structure but has not yet internalized it.
Beyond accuracy, \ourmethod also turns latent evidence use into readable
decisions---selection, indexed references, and an ordered chain---which give
compact models explicit process-level learning targets. For
Qwen2.5-VL-7B, training then further raises SlideVQA accuracy to $77.16$
(row \ourmethod, SlideVQA Overall in Table~\ref{tab:main}),
narrowing the gap to larger backbones.



\section{Conclusion}
We presented \ourmethod, a unified agentic framework for VRAG that renders evidence use explicit and controllable. 
At its core, a single retrieve--reason loop filters observations at collection time, then re-selects and reorders the retained evidence prior to answer generation, ensuring that every reasoning step is grounded in a traceable link to its supporting image. 
This design controls answer-context noise, and gains further improvements from filtered cold-start distillation and an evidence-selection RL reward that jointly teach evidence organization across benchmarks and backbones.

\bibliography{aaai2027}

\clearpage
\appendix
\section{Details of Search Engine}
Retrieval is backed by ColQwen2.5-v0.1~\cite{faysse2024colpali}. All page
images are encoded offline into patch-level multi-vector representations and
cached, so that at inference only the query needs to be embedded. The agent's
textual query is passed through the same encoder and scored against every
cached page via the late-interaction MaxSim operator, yielding a ranked list of
candidate pages. Rather than always returning the single top hit, the
environment keeps a per-trajectory record of pages already shown and returns the
highest-ranked \emph{unseen} page as the observation, which prevents the agent
from repeatedly inspecting the same evidence across turns. When every candidate
in the top-$k$ list has already been surfaced in previous turns, the environment
signals that no further pages are available and steers the agent toward emitting
its final answer.

\section{Necessity of Crop-and-Zoom Tool.}
\paragraph{Why cropping is needed.}
Following VRAG-RL~\cite{wang2026vrag} and VISOR~\cite{shen2026visor}, we bound the
input resolution of every page through a fixed \texttt{max\_pixels} cap instead
of feeding pages at native resolution. This keeps the visual-token cost per image
low, but it also means that once a full page is downsampled to that budget, its
dense local content---small charts, fine print, and tightly packed
tables---becomes illegible. Crop-and-zoom resolves this tension as a
\emph{dynamic resolution} control: it re-reads only the sub-region the agent
actually needs at high fidelity, while leaving the global token budget untouched.
Without it, the agent would be forced to choose between spending its entire budget
on a single high-resolution page and reasoning over unreadable compressed images;
cropping avoids both, trading one extra turn for exactly the resolution the
question demands.

\paragraph{Implementation.}
The \texttt{<bbox> [x1, y1, x2, y2] </bbox>} action carries pixel coordinates in the
coordinate frame of the image as shown to the VLM, which we first map through a
linear transform back onto the original high-resolution page. We then pad the box
by a fixed $28$-pixel margin on all sides, so that content adjacent to the region
is not cut off, and clamp the padded box to the page boundary to avoid
out-of-range indices. The resulting region is cropped and rescaled to a standard
resolution, letting the agent resolve fine-grained material---dense tables, small
charts, and the like---that is illegible at full-page scale. Malformed or
degenerate coordinates trigger an error message that asks the model to reissue the
action.

\begin{figure}[htbp]
\centering
\includegraphics[width=\columnwidth]{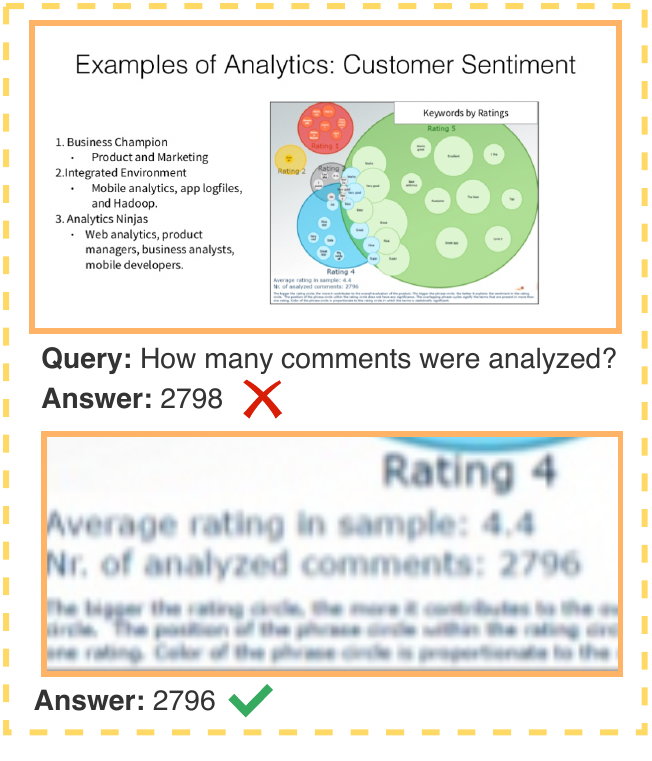}
\caption{A crop-and-zoom case study. At full-page resolution the answer-bearing
line is illegible and the agent misreads it as $2798$; after zooming into the
bounding box, the region is legible and the agent recovers the correct answer
$2796$. The error is perceptual, not a reasoning failure.}
\label{fig:crop_case}
\end{figure}

\paragraph{Case study.}
Figure~\ref{fig:crop_case} shows a concrete instance. The query \emph{``How many
comments were analyzed?''} is answered by a single line of fine print---``Nr.\ of
analyzed comments: 2796''---tucked beside a dense bubble chart on an otherwise
busy slide. At full-page resolution this line is blurred past the point of
legibility, and the agent misreads it as $2798$, an answer that is plausible and
close but wrong. Issuing a \texttt{<bbox>} over that region and re-reading it at
high fidelity makes the digits sharp, and the agent recovers the correct answer
$2796$. The error here is purely perceptual rather than a reasoning failure: the
evidence was on the retrieved page all along, and only the resolution to read it
was missing---exactly the gap crop-and-zoom is designed to close.

\section{Agent Loop Context}
\label{app:agent_loop}

\begin{figure*}[htbp]
\centering
\includegraphics[width=\textwidth]{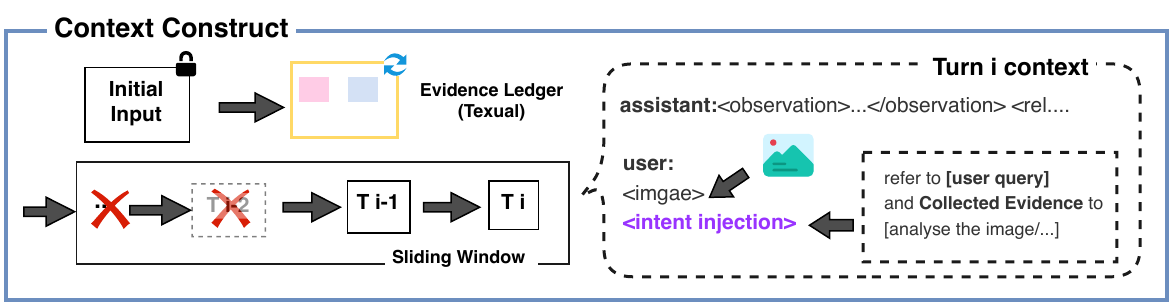}
\caption{Agent loop context construction. The persistent textual evidence ledger
(top) retains the distilled, query-relevant content of every past turn, while a
sliding window (bottom left) keeps only the most recent $W{=}2$ raw visual turns
and evicts older ones. Each returned observation carries an action-specific
intent-injection prompt (right) that restates the query and the collected
evidence and tells the agent what to do at that step---judging relevance after a
\texttt{search}, reading the crop after a \texttt{bbox}, reorganizing evidence
after \texttt{consolidate}, and answering from the consolidated evidence---so the
agent stays anchored to both the query and the current sub-goal.}
\label{fig:agent_loop}
\end{figure*}

This section expands on the context-construction rule of
Eq.~\eqref{eq:context} in the main text and illustrates it in
Figure~\ref{fig:agent_loop}. The difficulty it addresses is specific to
multi-image agentic VRAG: a full page image costs a large number of visual
tokens, so na\"ively concatenating every retrieved and zoomed image across a
long trajectory exhausts the context budget within a few turns and, worse, lets
early irrelevant images crowd out the query and induce semantic drift---the agent
gradually loses track of what it was asked. Following VISOR~\cite{shen2026visor},
we resolve this with two coupled mechanisms operating on a reconstructed
per-turn context rather than an ever-growing transcript.

\paragraph{Persistent ledger plus sliding window.}
At turn $t$ the context is rebuilt as
\begin{equation}
\mathbf{C}_t = \bigl[\,\mathbf{P}_{\text{init}};\;\mathcal{L}_t;\;
(\mathbf{r}_{t-W+1},o_{t-W+1}),\dots,(\mathbf{r}_t,o_t)\bigr],
\end{equation}
i.e.\ the system prompt and query $\mathbf{P}_{\text{init}}$, followed
immediately by the textual evidence ledger $\mathcal{L}_t$, followed by only the
most recent $W$ raw visual turns. The ledger is pinned right after the prompt so
that every relevance decision made so far stays visible in compact textual form,
decoupling \emph{what evidence has survived} from \emph{how many raw images are
currently in context}. The sliding window then caps the raw-image footprint at a
constant $W$ turns regardless of trajectory length: as shown on the left of
Figure~\ref{fig:agent_loop}, turns older than the window are evicted (crossed
out), but nothing is truly lost---their query-relevant content has already been
distilled into $\mathcal{L}_t$. We set $W{=}2$ because a single
\emph{retrieve-then-zoom} interaction spans two turns (a \texttt{search} that
returns a page, then a \texttt{bbox} that crops it); keeping the last two raw
turns preserves that chain intact while still bounding the image cost.

\paragraph{Intent injection.}
Even with a bounded window, over a long horizon the sheer volume of intermediate
observations can pull the agent away from the original question. To counter this,
every system-returned observation is augmented with an \textbf{intent injection}
prompt that restates the query $q$ and points back to the collected evidence, as
depicted in the \emph{Turn $i$ context} panel on the right of
Figure~\ref{fig:agent_loop}: the user turn pairs the returned image with an
instruction to ``refer to the user query and the collected evidence'' when
interpreting it. This re-anchors each perception step to the actual objective, so
that reading a new page is always framed by what the answer requires rather than
by whatever the last few turns happened to surface.

Crucially, the injected prompt is \emph{action-specific}: rather than a single
fixed reminder, each type of system return carries a tailored instruction that
tells the agent what to do next given the query and the evidence gathered so far.
When a \texttt{search} returns a new page, the prompt asks the agent to judge the
page's relevance to $q$ and, if relevant, distill its query-pertinent content into
the ledger. When a \texttt{bbox} returns a zoomed-in crop, the prompt directs the
agent to read the fine-grained region against $q$ and update the corresponding
ledger entry with what the magnified view reveals. When \texttt{consolidate}
returns the reloaded original pages, the prompt steers the agent to re-examine
them jointly---filtering residual noise and ordering the evidence into a coherent
chain---rather than treating them as yet another page to explore. Finally, at the
\texttt{answer} step the prompt re-states $q$ alongside the consolidated evidence
and asks the agent to produce the response strictly from it. Tying the injected
intent to the acting stage keeps every turn aligned not only with the original
query but with the specific sub-goal of that turn, so the agent neither drifts
from the question nor blurs the distinct roles of exploring, consolidating, and
answering.

\section{Evaluation Details}
\label{app:eval_details}

\begin{figure*}[htbp]
\centering
\includegraphics[width=\textwidth]{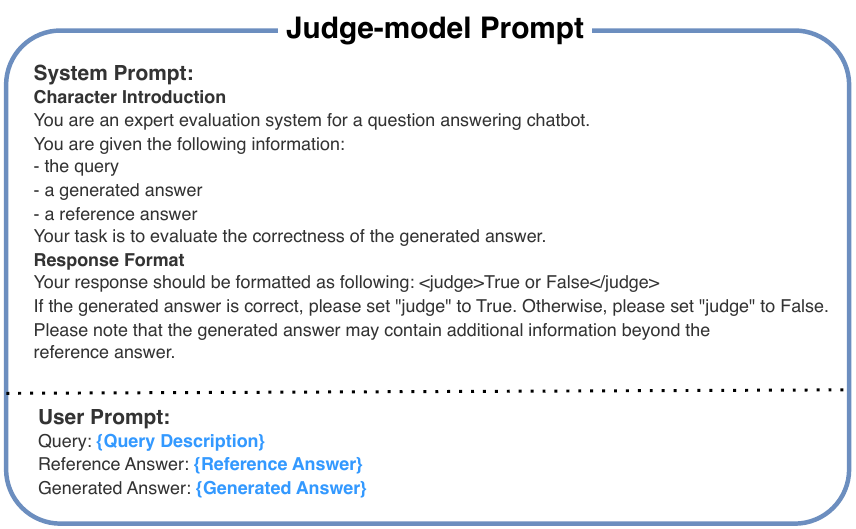}
\caption{The prompt template used for LLM-as-Judge evaluation.}
\label{fig:judge_prompt}
\end{figure*}

\paragraph{Judge Prompt.}
Following VRAG-RL~\cite{wang2026vrag} and VISOR~\cite{shen2026visor}, we use
\texttt{Qwen-max-latest} as the LLM judge to score model responses. Given the
question, the reference answer, and the model prediction, the judge returns a
binary label ($0$ or $1$) indicating whether the prediction is semantically
correct. The exact template is shown in Figure~\ref{fig:judge_prompt}.

\paragraph{Reliability of Model-as-Judge.}
Judging is reliable here because the task itself is objective: every query has a
ground-truth answer that is a short, factual phrase---a year, an institution, a
numeric value---so scoring reduces to checking a prediction against a fixed
reference rather than making an open-ended quality judgment. The reliability of
\texttt{Qwen-max-latest} in this role has already been validated by
VRAG-RL~\cite{wang2026vrag} and VISOR~\cite{shen2026visor}. To further corroborate
it, we re-evaluate our SlideVQA results with \texttt{DeepSeek-V3.2}~\cite{liu2024deepseek}
as an alternative judge. As shown in Table~\ref{tab:judge_compare}, the two judges
produce highly consistent scores overall, differing only slightly at the subtask
level. The small gap confirms that our reported results are not sensitive to the
choice of judge model.

\begin{table}[htbp]
\centering
\renewcommand{\arraystretch}{1.3}
\resizebox{\columnwidth}{!}{%
\begin{tabular}{lccc}
\hline
\textbf{Judge} & \textbf{Single-hop} & \textbf{Multi-hop} & \textbf{Overall} \\
\hline
Qwen-max-latest  & 82.28 & 62.26 & 77.16 \\
DeepSeek-V3.2      & 81.55 & 63.67 & 76.98 \\
\hline
\end{tabular}%
}
\caption{Comparison of SCORE scores on SlideVQA under two judge models (Qwen2.5-VL-7B).}
\label{tab:judge_compare}
\end{table}

\section{Dataset Information}\label{app:datasets}
Our evaluation spans three visually rich document benchmarks, each stressing a
distinct evidence regime.

\paragraph{SlideVQA.}
SlideVQA~\cite{tanaka2023slidevqa} poses questions over presentation slides drawn
from a broad mix of real-world decks and topics. Its questions fall into a
\textbf{Single-hop} subset ($1{,}648$) answerable from one slide and a
\textbf{Multi-hop} subset ($567$) that must aggregate evidence across several
slides of the same deck. We evaluate on the complete test split of $2{,}215$
questions.

\paragraph{ViDoSeek.}
ViDoSeek~\cite{wang2025vidorag} targets retrieval-augmented QA over a large
visually rich corpus of roughly $6{,}000$ page images mixing text, tables,
charts, and figures. Unlike SlideVQA and MMLongBench, which mix single- and
multi-hop questions, every ViDoSeek question is answerable from a single page:
the evidence for each answer lives on one image, so the challenge is locating
that page rather than aggregating across pages. Its $1{,}142$ test questions form
two disjoint groups by type: \textbf{Extraction} ($645$), which asks the model to
locate and read off a specific piece of information from the retrieved page, and
\textbf{Logic} ($497$), which additionally requires inference or computation over
that content to reach the answer.

\paragraph{MMLongBench.}
MMLongBench~\cite{ma2024mmlongbench} emphasizes long-context perception over
heterogeneous document content. Keeping only questions with verifiable reference
answers leaves $847$ evaluation questions, each tagged by content
type---\textbf{Text} ($291$), \textbf{Table} ($217$), \textbf{Chart} ($178$),
\textbf{Figure} ($290$), and \textbf{Layout} ($118$). A single question may carry
more than one tag, so these subsets overlap and their counts exceed $847$.

\section{Baseline Implementation Details}
\label{app:baselines}

Baseline numbers come from two sources. Results for Vanilla RAG, ReAct,
Search-R1-VL, VRAG-RL, and VISOR are quoted directly from their original
papers~\cite{faysse2024colpali,yao2022react,jin2025search,wang2026vrag,shen2026visor},
as our evaluation follows exactly the same experimental protocol, so re-running
them would reproduce the reported figures. M3RAG~\cite{du2026m3rag} is likewise
quoted from its paper, but under a setup that differs from ours in retrieval
configuration, which we flag alongside its result. The remaining four
baselines---ViDoRAG, EVisRAG, MMSearch-R1, and R1-Router---are reproduced by us
within our unified evaluation framework, using ColQwen2.5-v0.1~\cite{faysse2024colpali}
as the shared retriever so that only the agent design varies.

\paragraph{Vanilla RAG.}
A single-pass retrieval-augmented baseline: the original question retrieves
relevant pages, which are handed to the model for direct answer generation with
no iterative reasoning or multi-turn interaction. We adopt the visual variant, in
which page images are retrieved by ColQwen2.5-v0.1~\cite{faysse2024colpali} and
fed straight to the VLM.

\paragraph{ReAct.}
ReAct~\cite{yao2022react} casts the agent as an interleaved
Thought--Action--Observation loop for multi-turn retrieval-augmented reasoning.
Each turn issues a search query conditioned on the current reasoning state and
receives one retrieved page image as the observation, iterating until a final
answer is produced.

\paragraph{Search-R1-VL.}
A visual extension of Search-R1~\cite{jin2025search}, which brings multi-turn
RL-based reasoning into the RAG loop. The visual variant retargets this framework
to image-based retrieval, trained on the same data with the same reward and
post-processing as VRAG-RL and initialized from a cold-start checkpoint.

\paragraph{VRAG-RL.}
VRAG-RL~\cite{wang2026vrag} trains an agentic VLM with GRPO-based RL to iteratively
retrieve and reason over page images. It adds a crop-and-zoom tool for
fine-grained perception and uses trajectory-level rewards that jointly optimize
retrieval and answer quality.

\paragraph{VISOR.}
VISOR~\cite{shen2026visor} is a single-agent method that maintains a textual
evidence ledger over the interaction, recording query-relevant visual
observations as text to suppress noise from irrelevant images across long
horizons. As it shares our exact evaluation protocol, we quote its numbers
directly from the original paper.

\paragraph{M3RAG.}
M3RAG~\cite{du2026m3rag} is a multi-agent framework that splits the
retrieval--reasoning pipeline into specialized agents for multi-modal document
understanding. As its code has not been released, we report its numbers directly from the original paper.

\paragraph{ViDoRAG.}
ViDoRAG~\cite{wang2025vidorag} uses an actor--critic multi-agent architecture with
separate planning, retrieval, and answering agents for iterative reasoning over
visually rich documents. We plug in ColQwen2.5-v0.1~\cite{faysse2024colpali} as the
single-modal search engine, retrieving the top-$10$ pages per query, with
Qwen2.5-VL-7B~\cite{bai2025qwen2} as the backbone VLM; all other components follow
the original pipeline unchanged.
 
\paragraph{EVisRAG.}
EVisRAG~\cite{sun2025visrag} performs evidence-based reasoning over multiple
retrieved images, explicitly extracting and structuring per-page evidence to
support multi-image understanding. We substitute ColQwen2.5-v0.1~\cite{faysse2024colpali}
for the original VisRAG-Ret retriever, retrieving the top-$3$ pages per query, and
use the officially released EVisRAG-7B weights; all other settings match the
original configuration.

\paragraph{MMSearch-R1.}
MMSearch-R1~\cite{wu2025mmsearch} folds multimodal search into the reasoning loop
via cross-modal retrieval over both visual and textual forms. It ships two tools,
text search and image-to-image search; in our setting we adapt text search to
retrieve document-page images through ColQwen2.5-v0.1~\cite{faysse2024colpali},
while image search---inapplicable to document retrieval---returns a prompt
redirecting the model to text search. We use the officially released
MMSearch-R1-7B weights, with all other settings as in the original.

\paragraph{R1-Router.}
R1-Router~\cite{peng2025learning} uses a dynamic routing mechanism trained with
Step-GRPO: it generates intermediate sub-queries during reasoning and dispatches
each to the most suitable retrieval tool, curbing unnecessary retrievals while
adaptively integrating external evidence. In our setting every retrieval tool is
adapted to fetch document-page images via ColQwen2.5-v0.1~\cite{faysse2024colpali},
returning the top-$5$ pages per query, and the interaction budget is capped at $3$
turns; all other settings follow the official configuration.

\paragraph{Compute.}
All experiments run on $8\times$ NVIDIA A800 80GB GPUs.

\section{Training Hyperparameters}\label{app:hyperparams}

\begin{table}[h]
\centering
\renewcommand{\arraystretch}{1.3}
\begin{tabular}{lc}
\hline
\textbf{Name} & \textbf{Value} \\
\hline
Finetuning type              & Full    \\
Freeze vision tower          & True    \\
Freeze multi-modal projector & True    \\
Freeze language model        & False   \\
Cutoff length                & 16384   \\
Epochs                       & 3       \\
Batch size                   & 16      \\
Gradient accumulation steps  & 2       \\
Learning rate                & 1.0e-5  \\
LR scheduler type            & cosine  \\
Warmup ratio                 & 0.1     \\
\hline
\end{tabular}
\caption{Key hyperparameters for cold-start SFT (shared by the 7B and 3B backbones).}
\label{tab:hyper_sft}
\end{table}

\begin{table}[h]
\centering
\renewcommand{\arraystretch}{1.3}
\begin{tabular}{lc}
\hline
\textbf{Name} & \textbf{Value} \\
\hline
Number of agent groups       & 5       \\
Warmup steps ratio           & 0.285   \\
Train batch size             & 8       \\
Mini batch size per GPU      & 1       \\
Micro batch size per GPU     & 1       \\
Learning rate (Actor)        & 1.0e-6  \\
KL loss coefficient          & 0.01    \\
Tensor model parallel size   & 2       \\
Max prompt length            & 8192    \\
Max response length          & 2048    \\
Max turns                    & 10      \\
Total steps                  & 100     \\
GPU memory utilization       & 0.4     \\
\hline
\end{tabular}
\caption{Key hyperparameters for RL (shared by the 7B and 3B backbones).}
\label{tab:hyper_rl}
\end{table}

Table~\ref{tab:hyper_sft} and~\ref{tab:hyper_rl} list the full
hyperparameter settings for the cold-start SFT stage and the RL stage,
respectively. We deliberately keep a single configuration across both backbones:
the Qwen2.5-VL-7B and 3B models are trained under identical hyperparameters, so
that any performance difference between them reflects backbone capacity rather
than tuning. All runs use one node of $8\times$ NVIDIA A800 GPUs. The settings are
otherwise not tuned per benchmark; when porting SCORE to substantially larger or
smaller VLMs, scaling learning rate and context length accordingly is likely to
help.

\section{Details of Cold-Start}

We build the cold-start corpus with an automatic teacher-rollout pipeline on
SlideVQA training queries. Because the teacher follows the same SCORE action
protocol and the same ledger-plus-sliding-window context used at inference, we do
not restate those mechanics here (see the Method section and the Agent Loop
Context section above) and instead focus on what is specific to \emph{generating
and filtering} the data.

\paragraph{Teacher Rollout.}
For each query we prompt a stronger teacher, Qwen3.5-122B-A10B, to act as a SCORE
agent and interact with the retrieval environment under the fixed action schema
\texttt{observe},\texttt{relevant},\{\texttt{search}, 
\texttt{bbox}, \texttt{consolidate}\} defined in the main text: at each step it
reads the current image, marks whether that image carries query-relevant
evidence, and emits exactly one next action---a refined \texttt{search} when
evidence is still insufficient, a normalized \texttt{bbox} when a relevant page
hides answer-critical local detail (table cells, chart values, axes, small text,
names, dates, or numbers), or \texttt{consolidate} once enough evidence is in
hand. The environment logs both the teacher messages and its returned visual
observations: a \texttt{search} appends the retrieved page image, and a
\texttt{bbox} appends the cropped region as the next observation. To bound visual
token cost, every image handed to the teacher is resized to a fixed pixel budget,
constrained between $256\times28\times28$ and $512\times28\times28$. Each
trajectory is capped at a maximum number of interaction steps; if the teacher does
not consolidate within that budget, the environment forces consolidation over the
evidence collected so far and proceeds to answer generation. The resulting
trajectory thus records the complete trace: per-turn structured reasoning,
retrieval actions, optional region zooming, the consolidation index list, and the
final answer.

\paragraph{Trajectory Filtering.}
Teacher rollouts can still contain noisy or shortcut solutions---most dangerously,
answers judged correct without actually visiting all the evidence they should
depend on. We therefore keep a trajectory only if it passes two conjunctive
checks. \emph{(i) Answer correctness.} An LLM judge receives the question, the
reference answer, and the teacher's final answer and returns a binary label; it is
instructed to accept semantic, numerical, and abbreviation equivalences (e.g.,
treating ``2Bn'' and ``2 billion'' as the same). \emph{(ii) Full evidence
coverage.} From each SlideVQA sample we build the set of gold reference pages
$\mathcal{P}_{\mathrm{gold}}$ (from the source file name and the annotated
reference indices) and the set of pages retained in the trajectory's evidence
ledger $\mathcal{P}_{\mathrm{ledger}}$---i.e., the source pages of the
observations the teacher marked \texttt{relevant}=\texttt{yes} and kept---and
require
\[
\mathrm{JudgeCorrect}(a, a^\star)=1
\quad\land\quad
\mathcal{P}_{\mathrm{gold}}\subseteq\mathcal{P}_{\mathrm{ledger}}.
\]
Samples lacking reference-page metadata are exempt from the coverage test rather
than discarded. This filter is deliberately conservative: the judge removes
wrong-answer trajectories, while the coverage constraint removes false
positives---trajectories that match the answer without \emph{retaining} every
annotated page as evidence---so the student is less likely to imitate spurious
retrieval or incomplete evidence selection. Note that coverage is checked against
the kept ledger rather than every page the teacher merely glanced at: a gold page
that was retrieved but discarded as \texttt{relevant}=\texttt{no} does not count,
which is exactly what makes the criterion supervise \emph{selection} and not just
retrieval.

\paragraph{SFT Conversion and Supervision.}
Each surviving trajectory is converted into a multimodal chat-style SFT instance:
text is kept verbatim, while every environment-provided image is replaced by an
\texttt{<image>} placeholder aligned to a separate image list. We drop malformed
instances---empty assistant turns, trajectories whose final message is not the
assistant's, and those exceeding a fixed image budget---while preserving the full
agentic format (initial instruction, per-turn teacher outputs, returned or cropped
images, consolidation output, and final answer). For training simplicity,
cold-start supervises the \emph{complete} trajectory: unlike inference and RL,
where the context is reconstructed per turn with only the ledger plus the last $W$
raw visual turns, the SFT instance keeps every returned image in place, with
neither a sliding window nor a dynamically rebuilt ledger injected into the
context. What the student is explicitly supervised to reproduce is therefore three
behaviors along the raw trace: (i) the multi-turn \texttt{search}/\texttt{observe}
retrieval-and-reading loop; (ii) the per-image \texttt{relevant}=\texttt{yes/no}
decision that filters evidence at collection time; and (iii) at
\texttt{consolidate}, the rerank over the evidence accumulated up to that point
together with the final answer generated from it. Learning these on the full
rollout lets the student internalize the ledger and consolidation behavior before
the sliding-window context is imposed at deployment. Training uses assistant-only
supervision: loss is computed solely on the teacher's structured generations
(\texttt{observe}, \texttt{relevant}, action decisions, consolidation indices, and
answer), with user instructions and environment observations serving as context
only. This stage teaches the student the SCORE protocol, ledger behavior, and
consolidation format, providing a reliable initialization before RL further
sharpens query-aware evidence selection.

\section{On an Evidence Re-Fetch Module}

\paragraph{Motivation.}
Retrieval order in a multi-hop query is not deterministic, and this raises a
concern about our collection-time relevance judgment. Consider a question such as
\emph{``for the company ranked third by metric A in 2010, what is its metric B in
2014?''} The retriever may surface the 2014 metric-B page \emph{before} the 2010
ranking is known, so at the moment that page is inspected the agent cannot yet
verify whether it concerns the target company. A strict relevance filter might
then discard it as irrelevant and, because that discard is permanent, lose a gold
page it will later need. This suggests an \textbf{evidence re-fetch} module: after
consolidation, allow the agent to reconsider and pull back pages it had earlier
dropped, once the query constraints have become clear.

\paragraph{Finding.}
We implemented such a module and found that it does not help and in fact slightly
hurts (Table~\ref{tab:refetch}). Two observations explain why. First, the relevance
judgment rarely discards gold pages to begin with: SCORE already keeps a ledger
coverage of $78.4$ (main-text analysis), so the feared ``premature discard'' is
uncommon---when uncertain whether a page qualifies, the trained policy defaults to
\emph{keeping} it rather than dropping it, and the ledger errs toward
over-retention, exactly the safe direction for recall. Consistent with this,
adding re-fetch barely moves coverage at all ($78.4\rightarrow78.5$): there is
almost nothing left to recover. Second, and more decisively, the few pages that
\emph{are} dropped cannot be brought back usefully. Inspection shows these are not
order artifacts but genuine perception failures---the model misread the page's
content, so it would misuse the same page even if handed it back. Worse, letting
the agent reopen previously rejected pages reintroduces the very noise the
relevance filter was meant to remove, nudging a few borderline cases from correct
to wrong, so overall accuracy edges \emph{down} rather than up
($77.16\rightarrow76.79$).

\begin{table}[htbp]
\centering
\renewcommand{\arraystretch}{1.3}
\resizebox{0.8\columnwidth}{!}{%
\begin{tabular}{lcc}
\hline
\textbf{Variant} & \textbf{Ledger Cov.} & \textbf{Acc.} \\
\hline
SCORE                     & 78.4 & \textbf{77.16} \\
\quad + evidence re-fetch & 78.5 & 76.79 \\
\hline
\end{tabular}%
}
\caption{Effect of adding a post-consolidation evidence re-fetch module on
SlideVQA (Qwen2.5-VL-7B). Ledger Cov.\ is the fraction of gold pages kept in the
ledger. Re-fetch leaves coverage essentially unchanged---there is little to
recover---while slightly lowering accuracy by re-admitting previously rejected
pages.}
\label{tab:refetch}
\end{table}

\paragraph{Key findings.}
We therefore drop the re-fetch module. Under our current benchmarks the relevance
judgment already keeps coverage high, the pages it does drop are lost to
misperception rather than to retrieval order---so re-fetch recovers almost nothing
($+0.1$ coverage)---and reopening rejected pages only lets noise back in, slightly
lowering accuracy. We note, however, that these benchmarks may be relatively
benign for this question: longer horizons, more hops, or harder
constraint-ordering could make premature discards more frequent and a
well-designed re-fetch more valuable, and we leave that exploration to future
work.

\section{Main-Text Ablation Details}

\subsection{Training the Ablation Variants}
The main-text ablation (Table~\ref{tab:ablation}) reports each variant under
both an untrained \emph{Vanilla} setting and a \emph{Fine-tuned}
(cold-start\,+\,RL) setting. To avoid confounding architecture with training,
every \emph{Fine-tuned} variant is \textbf{retrained from scratch} under the
exact same pipeline as full SCORE---the same teacher, cold-start filtering, GRPO
configuration, and hyperparameters---rather than obtained by disabling a module
at inference time on the full model. The two structural designs, the
sliding-window context and the persistent evidence ledger, are retained in
\emph{all} variants; what changes is only which of the two agentic
\emph{actions}---relevance judgment and consolidation---the policy is allowed to
emit.

\paragraph{What each variant does.}
Concretely, all variants reload the original page images referenced by the ledger
before answering; they differ only in whether relevance judgment and evidence
consolidation are performed:
\begin{itemize}
\item \textbf{w/o relevance judgment.} The per-turn
  \texttt{<relevant> yes/no </relevant>} decision is removed, so each turn produces
  only an observation and the next action; every retrieved page is written into
  the ledger unfiltered. Before answering, the original images referenced by the
  ledger are reloaded and consolidation still re-selects and reorders them.
\item \textbf{w/o consolidation.} The global re-selection, denoising, and
  reordering step is removed. When the agent has gathered sufficient evidence, it
  reloads the original images referenced by the accumulated ledger and answers
  directly from all of them in their original collection order.
\item \textbf{w/o both.} Neither relevance judgment nor consolidation is
  available. Every retrieved page is written into the ledger unfiltered; once the
  agent judges the gathered information sufficient, it reloads all original page
  images referenced by the ledger and answers from them in their original
  collection order, without collection-time filtering or pre-answer organization.
\end{itemize}

\paragraph{Reward.}
All \emph{Fine-tuned} variants keep the full ledger-based reward of
Eq.~\ref{eq:reward}, including the coverage and compactness terms. We
deliberately do \emph{not} weaken the reward for the ablated variants: the
evidence-selection signal is applied identically in every case. For variants
without consolidation the reward is computed on the final accumulated ledger
(which coincides with the consolidated ledger when consolidation is absent), so
every variant is optimized toward the same evidence-coverage objective and any
accuracy gap reflects the missing action rather than a different training target.

\section{Additional Ablations}
Both ablations in this section are conducted on SlideVQA, the source of our
training data, and we report accuracy (\%) on its test set.

\paragraph{Persistent evidence ledger.}
The textual evidence ledger is a core mechanism of SCORE and cannot simply be
deleted, since consolidation, the final answer, and the relevance judgment all
build on it. What we test here is instead its role \emph{during iterative
retrieval}. This requires removing the sliding window at the same time: the ledger
and the raw-image window are SCORE's two carriers of cross-turn memory, so if we
dropped only the textual ledger, the recent images still visible in the window
would silently stand in for it and mask its effect. Removing both leaves the agent
with no explicit memory of what it has already found across turns---each step sees
only the current observation---which is exactly the condition that isolates the
ledger's contribution to long-horizon retrieval. The agent can still mark each
observed page as relevant or not, and consolidation still runs before answering.
However, without a running textual memory the retrieval process starts to
\emph{drift}: lacking a compact anchor of the evidence gathered so far, the agent
brings in more and more noise as the trajectory grows, and its successive queries
wander away from the target. Even though relevance filtering and final
consolidation are still present, this drift happens during collection, so in cases
that need several retrieval attempts the agent may never surface the key image at
all. As shown in Table~\ref{tab:ledger_ablation}, removing the iterative textual
ledger lowers final accuracy, confirming that maintaining evidence as running
text---not only consolidating it at the end---is what keeps long-horizon retrieval
on track.

\begin{table}[htbp]
\centering
\renewcommand{\arraystretch}{1.3}
\resizebox{0.7\columnwidth}{!}{%
\begin{tabular}{lc}
\hline
\textbf{Variant} & \textbf{Acc.} \\
\hline
SCORE & 77.16 \\
\quad w/o iterative ledger & 73.09 \\
\hline
\end{tabular}%
}
\caption{Effect of removing the iterative textual evidence ledger (together with
the sliding window).}
\label{tab:ledger_ablation}
\end{table}

\paragraph{Sliding-window context.}
As an additional study, we keep the textual evidence ledger and vary only the
raw-image window size, comparing $W{=}1,2,3$ against a \emph{w/o} setting that
keeps no raw image at all. A smaller window is cheaper but may drop the page
needed to interpret a recent crop; a larger window keeps more images but also
brings more visual tokens and irrelevant history into the prompt. As shown in
Table~\ref{tab:window_ablation}, $W{=}2$ gives the best trade-off: it preserves the
common retrieve--zoom pair while leaving older evidence to the ledger.

\begin{table}[htbp]
\centering
\renewcommand{\arraystretch}{1.3}
\resizebox{0.95\columnwidth}{!}{%
\begin{tabular}{lcccc}
\hline
\textbf{Metric} & \textbf{$W{=}1$} & \textbf{$W{=}2$} & \textbf{$W{=}3$} & \textbf{w/o} \\
\hline
Acc. & 76.66 & 77.16 & 76.16 & 69.21 \\
\hline
\end{tabular}%
}
\caption{Effect of the sliding-window size (textual ledger kept in all settings).}
\label{tab:window_ablation}
\end{table}

\section{Prompt Template}

\begin{figure*}[htbp]
\centering
\includegraphics[width=\textwidth]{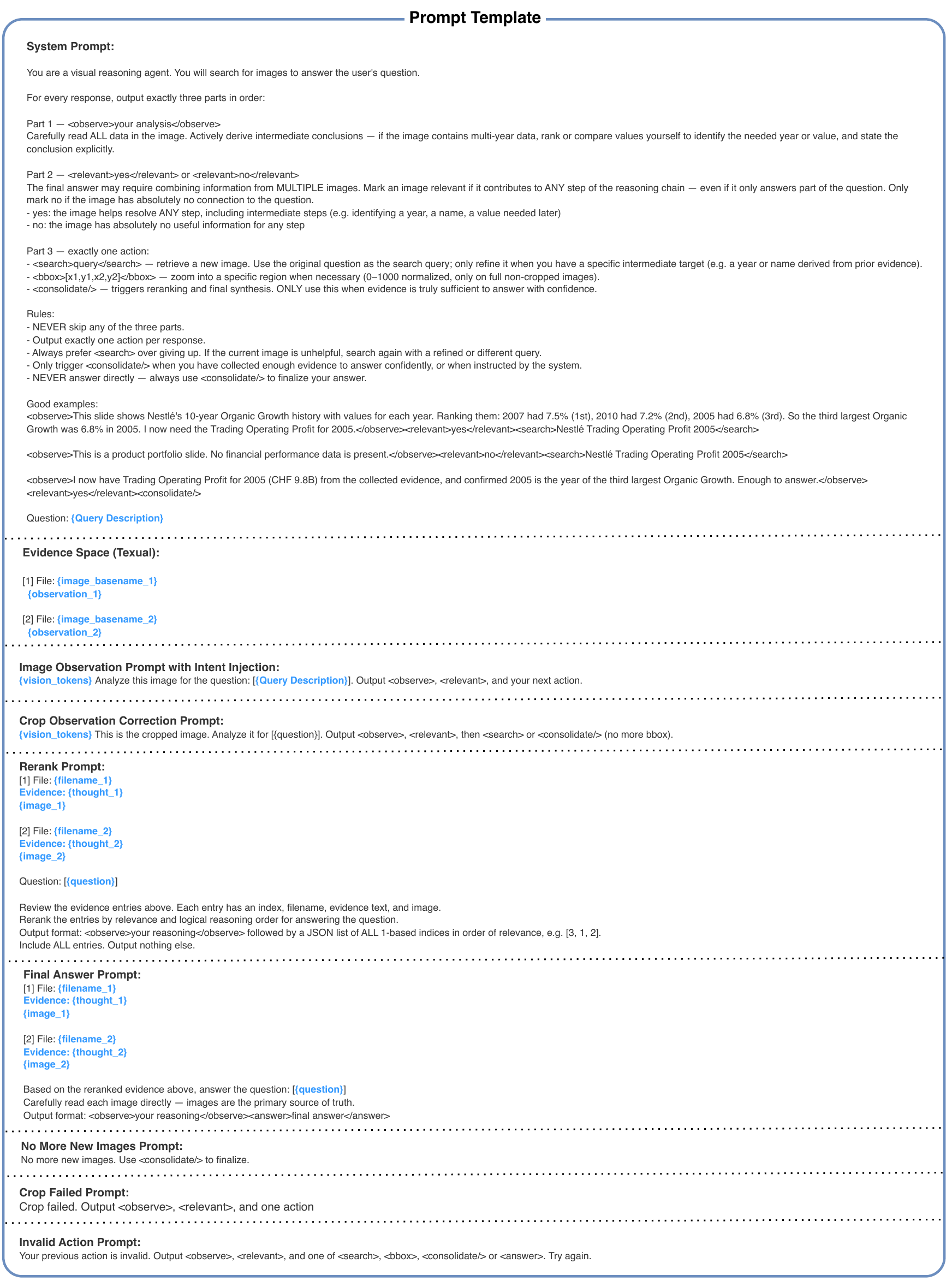}
\caption{Prompt template of SCORE. Blue fields are runtime variables inserted by
the environment.}
\label{fig:prompt_template}
\end{figure*}

Figure~\ref{fig:prompt_template} summarizes the prompt templates used by SCORE
throughout the agent loop. The template contains the system instruction, the
textual evidence space, action-specific observation prompts, the reranking prompt
for consolidation, and the final-answer prompt. Blue fields denote runtime
variables filled by the environment, such as the user question, returned visual
tokens, image file names, and evidence entries.

\section{Case Study}

\begin{figure*}[htbp]
\centering
\includegraphics[width=\textwidth]{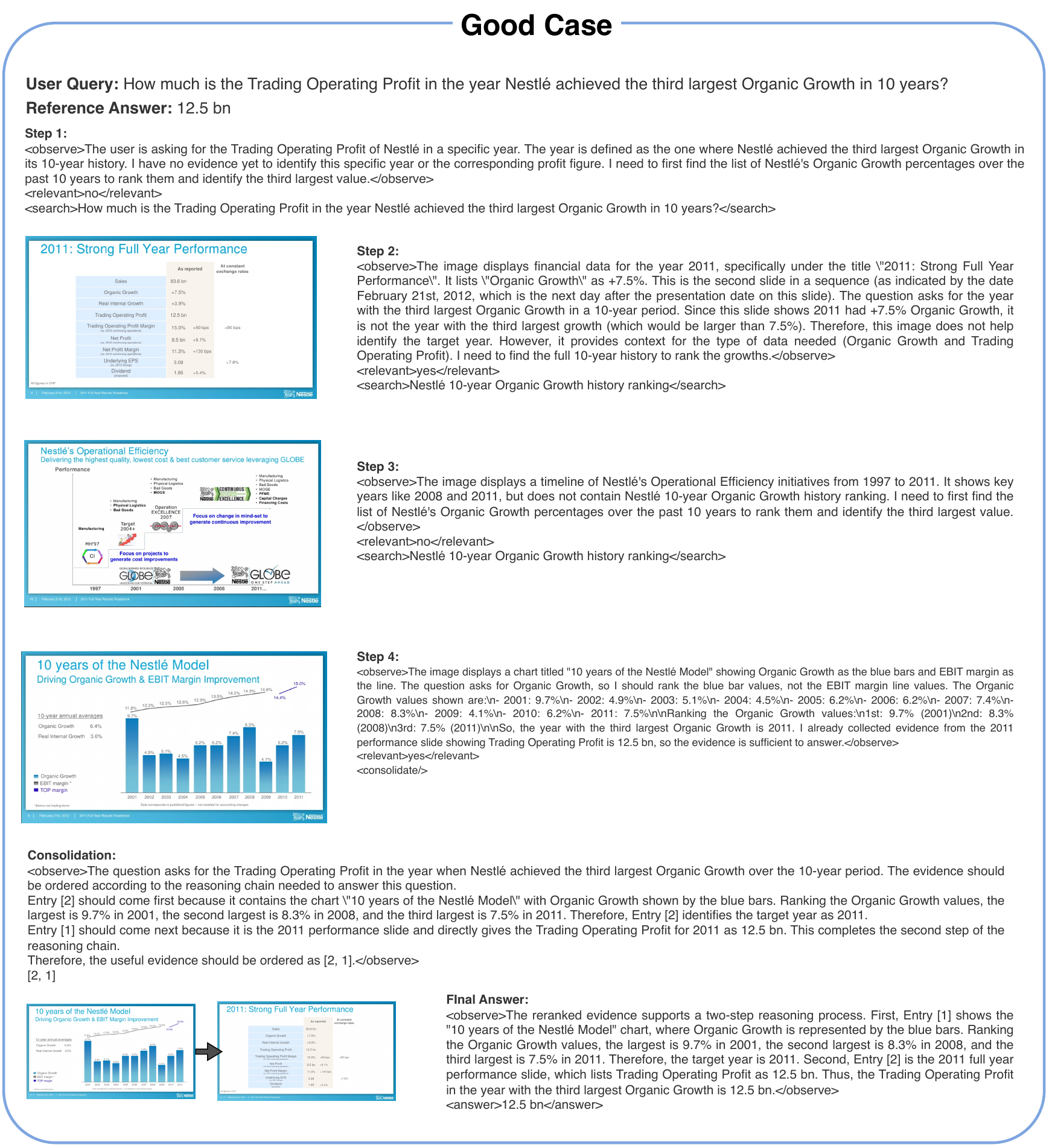}
\caption{A good case showing relevance filtering and evidence consolidation.}
\label{fig:good_case}
\end{figure*}

\paragraph{Good case.}
Figure~\ref{fig:good_case} shows a representative multi-hop example where SCORE's
relevance filtering and consolidation work together. The question asks for
Nestl\'{e}'s Trading Operating Profit in the year with the third-largest Organic
Growth over a ten-year period. During collection, SCORE does not treat every
retrieved Nestl\'{e} slide as useful: it rejects an operational-efficiency slide
that does not contain either the growth ranking or the profit value, while keeping
the financial-performance slide and the ten-year growth chart. After the growth
chart identifies 2011 as the target year, consolidation reorders the retained
evidence into a readable chain: first the ranking page that establishes the year,
then the 2011 financial page that gives the profit. The final evidence table and
answer therefore expose the reasoning path directly, rather than leaving the
reader to inspect the full retrieval history.

\begin{figure*}[htbp]
\centering
\includegraphics[width=\textwidth]{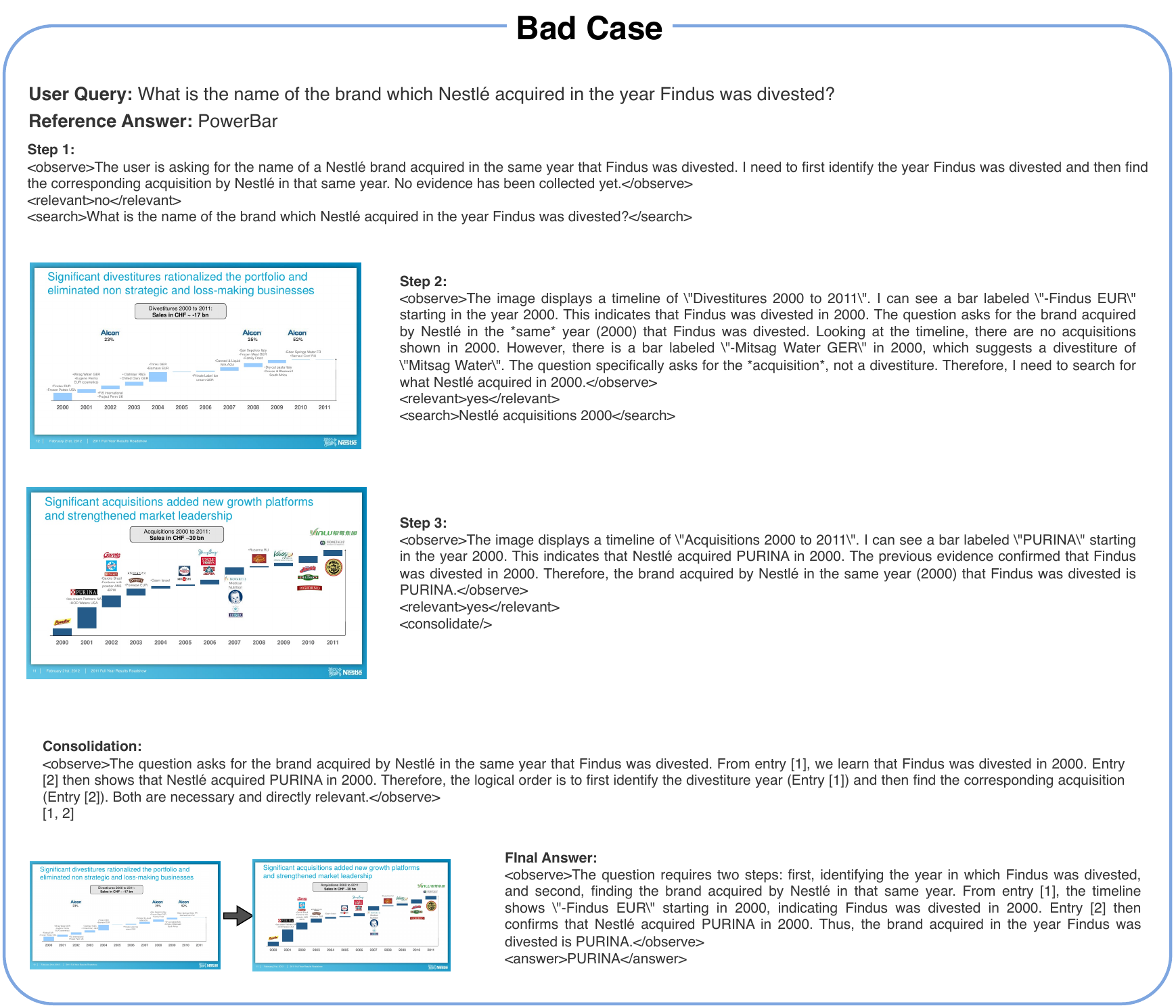}
\caption{A bad case where the correct image is retrieved but chart-structure
understanding leads to the wrong answer.}
\label{fig:bad_case}
\end{figure*}

\paragraph{Bad case.}
Figure~\ref{fig:bad_case} shows a failure mode where retrieval succeeds but visual
structure understanding fails. The agent correctly retrieves the divestiture
slide and identifies that Findus was divested in 2000. It also retrieves the
right acquisition slide for the same year. However, the acquisition chart places
multiple brand logos around the 2000 bar, including PowerBar and Purina, and the
model incorrectly associates the year with Purina instead of the target brand
PowerBar. Thus the error is not caused by missing evidence or search failure: the
answer-bearing image is already present in the trajectory, but the model fails to
parse the chart layout and bind the correct logo to the year. This case suggests
that SCORE's evidence selection and consolidation can expose the right pages, yet
fine-grained structural perception inside a dense visual chart remains a limiting
factor.

\end{document}